\documentclass[sigconf]{acmart}

\usepackage{booktabs} 
\usepackage{algorithm}
\usepackage{amsmath}
\usepackage[noend]{algpseudocode}
\usepackage{multirow}
\usepackage{makecell}
\usepackage{scalerel}
\usepackage{balance}

\copyrightyear{2024} 
\acmYear{2024} 
\setcopyright{acmlicensed}\acmConference[KDD '24]{Proceedings of the 30th ACM SIGKDD Conference on Knowledge Discovery and Data Mining}{August 25--29, 2024}{Barcelona, Spain}
\acmBooktitle{Proceedings of the 30th ACM SIGKDD Conference on Knowledge Discovery and Data Mining (KDD '24), August 25--29, 2024, Barcelona, Spain}
\acmDOI{10.1145/3637528.3671821}
\acmISBN{979-8-4007-0490-1/24/08}

\usepackage{tabulary}
\newcolumntype{C}[1]{>{\centering\arraybackslash}p{#1}}
\usepackage{bm}

\usepackage{setspace}
\usepackage{enumitem}
\setlist{nolistsep}

\begin{document}

\title{Hate Speech Detection with Generalizable Target-aware Fairness}

\author{Tong Chen}
\affiliation{%
  \institution{The University of Queensland}
  \country{Brisbane, Australia}
}
\email{tong.chen@uq.edu.au}

\author{Danny Wang}
\affiliation{%
  \institution{The University of Queensland}
  \country{Brisbane, Australia}
}
\email{danny.wang@uq.edu.au}

\author{Xurong Liang}
\affiliation{%
  \institution{The University of Queensland}
  \country{Brisbane, Australia}
}
\email{xurong.liang@uq.edu.au}

\author{Marten Risius}
\affiliation{%
  \institution{The University of Queensland}
  \country{Brisbane, Australia}
}
\email{m.risius@business.uq.edu.au}

\author{Gianluca Demartini}
\affiliation{%
  \institution{The University of Queensland}
  \country{Brisbane, Australia}
}
\email{g.demartini@uq.edu.au}

\author{Hongzhi Yin}
\authornote{Corresponding author.}
\affiliation{%
  \institution{The University of Queensland}
  \country{Brisbane, Australia}
}
\email{h.yin1@uq.edu.au}

\renewcommand{\shortauthors}{Tong Chen et al.}

\begin{abstract}
To counter the side effect brought by the proliferation of social media platforms, hate speech detection (HSD) plays a vital role in halting the dissemination of toxic online posts at an early stage. However, given the ubiquitous topical communities on social media, a trained HSD classifier can easily become biased towards specific targeted groups (e.g., \textit{female} and \textit{black} people), where a high rate of either false positive or false negative results can significantly impair public trust in the fairness of content moderation mechanisms, and eventually harm the diversity of online society. Although existing fairness-aware HSD methods can smooth out some discrepancies across targeted groups, they are mostly specific to a narrow selection of targets that are assumed to be known and fixed. This inevitably prevents those methods from generalizing to real-world use cases where new targeted groups constantly emerge (e.g., new forums created on Reddit) over time. 
To tackle the defects of existing HSD practices, we propose \underline{Ge}neralizable \underline{t}arget-aware \underline{Fair}ness (GetFair), a new method for fairly classifying each post that contains diverse and even unseen targets during inference. To remove the HSD classifier's spurious dependence on target-related features, GetFair trains a series of filter functions in an adversarial pipeline, so as to deceive the discriminator that recovers the targeted group from filtered post embeddings. 
To maintain scalability and generalizability, we innovatively parameterize all filter functions via a hypernetwork. Taking a target's pretrained word embedding as input, the hypernetwork generates the weights used by each target-specific filter on-the-fly without storing dedicated filter parameters. In addition, a novel semantic gap alignment scheme is imposed on the generation process, such that the produced filter function for an unseen target is rectified by its semantic affinity with existing targets used for training. Finally, experiments\footnote{The implementation of GetFair is released at \url{https://github.com/xurong-liang/GetFair}.} are conducted on two benchmark HSD datasets, showing advantageous performance of GetFair on out-of-sample targets among baselines. 

\end{abstract}



\maketitle

\section{Introduction}\label{sec:intro}
Many benefits of social media's liberation of communication come at the expense of proliferating hate speech. To prevent the negative socio-economical impact from hateful content, accurate algorithms for hate speech detection (HSD) have been heavily investigated by both industry practitioners~\cite{dixon2018measuring} and research communities \cite{fortuna2018survey}. 

Meanwhile, on the flip side of the coin, it has been reported \cite{garg2023handling} that various HSD algorithms have exposed vulnerability to different types of biases, including identity, annotation, and political biases. In the context of HSD, these biases are predominately associated with the sources (i.e., authors) of online posts, and a variety of corresponding solutions have also been made available. However, there also exists bias towards the \textit{targets}, or more precisely the posts' \textit{targeted groups} \cite{garg2023handling}, which are usually an identity group (e.g., African Americans), or a particular protected user attribute (e.g., one's religion). In the rest of this paper, when there is no ambiguity, we will use target to refer to targeted groups in a post. 
The targets of a post can be identified based on the main topics discussed, the context of conversation, or the channels hosting this post (e.g., the \textit{incel} forum on Reddit). 
Due to the inherent disparity in label distributions and language styles among different targets, 
models trained on such skewed data can reflect highly unstable HSD performance across targets. As a result, it is commonly seen that HSD classifiers exhibit abnormally high false positive or negative rates on some targeted groups. A high false positive rate on a specific target means that, the HSD classifier is prone to misclassifying a neutral post as hateful as long as this target is mentioned, introducing a greater chance of seeing incorrectly blocked posts in social media applications. In contrast, the high rate of false negative predictions are a result of failing to recognize and moderate truly hateful content, which in turn makes the specific target group more vulnerable to being attacked online. 

As such, achieving target-aware fairness in HSD becomes an increasingly important requirement for today's content moderation practices, and an emerging area to be researched. The uneven target-wise HSD performance implies that a spurious correlation between specific targets and the labels \cite{lipton2018does,ramponi2022features} (e.g., \textit{hateful} and \textit{neutral} in a binary setting) is established by the classifier. The spurious correlation then misleads the HSD classifier to provide incorrect labels with a groundlessly high confidence. To prevent a classifier from such biases, the ultimate goal is to lower its sensitivity to spurious features like target-related texts (i.e., \textit{who} are being discussed), and instead base its judgement on more generalizable linguistic features (i.e., \textit{how} they are being discussed). In this way, balanced HSD performance across targets can be attained. 

To pursue this goal, relevant literature has seen various data-centric methods that append new data sources or alter the way of engaging data samples to debias HSD models. Examples include sample reweighting \cite{zhang2020demographics,schuster2019towards,zhou2021challenges} that weakens the weight of samples with spurious features and lays more emphasis on samples that are less prone to biases during training, as well as marking bias-sensitive words \cite{ramponi2022features,dixon2018measuring,kennedy2020contextualizing} to perform label corrections or regulate model-level fairness. Nevertheless, these solutions mostly come with a strong empirical nature, such as the need for identifying confounding lexical patterns beforehand \cite{schuster2019towards} and the labor-intensive process of manual annotation \cite{ramponi2022features}. This has stimulated the emergence of less empirical, model-centric fair HSD solutions that are often designed under the principle of ``fairness through unawareness''~\cite{mehrabi2021survey}. In a nutshell, filtering modules are designed as an add-on to the HSD model, where information related to protected attributes are implicitly removed from learned representations of an online post. Then, methods like regularization \cite{hauzenberger2023modular}, multi-task learning \cite{gupta2023same}, and adversarial training \cite{kumar2023parameter} are adopted to jointly optimize the information filters along with the HSD task. 

Despite the stronger performance of filter-based solutions and their plug-and-play compatibility with most HSD classifiers, their practicality is inevitably challenged by the diversity of targets in real-world HSD applications.
Generally, the majority of the target debiasing methods in HSD bear an assumption that the number and identity of targets are fixed and consistent (e.g., only \textit{race} and \textit{gender}) across both training and inference phases. 
However, due to the difficulty in obtaining high-quality labeled datasets for HSD~\cite{garg2023handling}, the targets of acquired posts in the training data are just a tip of the iceberg compared with the ones in the online environment, rendering this assumption ill-posed and hurting the generalizability of existing debiasing approaches. 
For most solutions, their filters are only designed and trained for targets that are known in the training stage, and are thus hardly transferrable to an unseen target group. 
Furthermore, a single filter function, commonly parameterized as a neural network such as multi-layer perceptron (MLP), is subject to limited capacity for removing multifaceted target-specific information from textual embeddings, bringing the need for target-specific content filters. Considering the amount of possible targets being discussed online, it is infeasible to learn a dedicated content filter per target from both scalability and data availability perspectives. Given the high velocity of evolving targets, and naturally emerging new targets of posts on social media platforms (e.g., new Reddit forums and Facebook hashtags created daily), the capability of generalizing to new targets without being constantly retrained is deemed crucial.

To this end, in this paper we propose a novel approach, namely HSD with \underline{Ge}neralizable \underline{t}arget-aware \underline{Fair}ness (GetFair). In GetFair, to avoid the linear parameter cost associated with the number of target-specific filter functions, instead of independently parameterizing each filter, we put forward a hypernetwork \cite{von2020continual} that adaptively generates filter parameters -- MLP weights and biases in our case -- for each target. When debiasing post embeddings with an identified target described by one or several keywords, the input to the hypernetwork is defined as the target's indicator embedding, which can be easily composed with off-the-shelf pretrained word embeddings\footnote{GetFair adopts GloVe \cite{pennington2014glove} word embeddings.}.  
On the one hand, as the target embedding serves as the indicator to control the parameter generation process, the uniqueness of generated target-specific filters is guaranteed. On the other hand, by ensembling target-specific filters, GetFair is able to efficiently handle cases where a combination of different targets are identified in the same post, which is a trickier yet understudied case in real-world scenarios \cite{lalor2022benchmarking}. 
Moreover, the readily available target embeddings makes it possible for the hypernetwork to generate filter parameters with a valid indicator for an arbitrary, unseen target. An adversarial training paradigm is in place to optimize the filters toward learning debiased post embeddings. Concretely, a target classifier is deployed as the discriminator and tries to infer the original targets from the filtered post embeddings, where a capable filter function is ultimately learned to maximize the classification error of the discriminator. To further refine the dynamic filter generation for each target that arrives, a semantic gap alignment scheme is proposed, such that the semantic distance among all targets' input embeddings is resembled in the parameter space of their corresponding filters. 

To sum up, the contributions of this paper are three-fold:
\begin{itemize}
	\item We focus on an emerging problem in HSD concerning prediction fairness across different targeted groups of online posts by isolating the HSD classifier's judgement from a post's target-related spurious features. We point out the generalization deficiency of existing target-aware HSD debiasing methods when facing newly identified targets that are unknown during training. 
	\item A remedy, namely GetFair is proposed to achieve generalizable target-aware fairness as a plug-and-play method for most HSD classifiers. Instead of training a filter function for each target, GetFair optimizes a hypernetwork that adaptively parameterizes target-specific filters based on the indicators it receives. The generated filters are adversarially trained and further regularized via an innovative semantic gap alignment constraint, uplifting their debiasing capability without the pressure on scalability. 
	\item We conduct extensive experiments on two public benchmark datasets, where the results have demonstrated the superiority of GetFair, evidenced by its balanced effectiveness-fairness trade-off in HSD tasks on out-of-sample targets compared with state-of-the-art HSD debiasing baselines. 
\end{itemize}

\section{Preliminaries}\label{sec:prelim}
In this section, we mathematically define the concept of target-aware fairness in HSD, metrics for quantifying such fairness, and our research objective w.r.t. generalizable target-aware HSD. 

\textbf{Hate Speech Detection Tasks.} Hate speech detection is commonly defined as a classification task. In our paper, the default task setting is to take a social media post $s$ consisting of a sequence of tokens as the input, and output a predicted scalar $\hat{y} \in (0,1)$ to represent the likelihood of $s$ being a hate speech. In the most recent literature \cite{mathew2021hatexplain,tekirouglu2020generating}, pretrained transformer models like BERT \cite{vaswani2017attention} and GPT \cite{radfordimproving} are commonly the default encoder $g(\cdot)$ for generating the post-level embeddings $\textbf{s}=g(s)$ given their superiority in representation learning, which is also the initial feature we feed into the HSD classifier. Note that our paper is scoped around this text-based, binary classification setting, which is the most widely used one \cite{garg2023handling} in HSD. Meanwhile, with the availability of more nuanced annotated data, our findings can be easily generalized to some emerging settings like multi-class classification with different levels of hatefulness \cite{zampieri2019predicting} or HSD with additional multimodal data (e.g., images \cite{kiela2020hateful}) given their text-focused origin. 

\textit{\textbf{Definition 1}: Targets in A Post.} In our setting, each social media post $s_i$ has one or more targeted user groups being discussed. In $s_i$, those mentioned user groups are termed \textit{targets}, denoted by $t\in \mathcal{T}_i$. If deployed in a real online environment, one can consider using rule- or lexical-based approaches \cite{silva2016analyzing,elsherief2018hate}, or trainable detectors \cite{sharma2023characterizing} to identify the target set $\mathcal{T}_i$ from each post $s_i$. Because our emphasis is to uplift target-aware HSD fairness, we directly employ datasets (see Section \ref{sec:data} for details) that come with target labels, which are identified by human annotators from the textual clues within each post \cite{mathew2021hatexplain,ousidhoum2019multilingual,sheth2024causality}. The target identification task, as a more widely studied task, is not investigated in the paper.

\textit{\textbf{Definition 2}: Target-aware Fairness Metrics.} We assume a finite set of targeted groups $\mathcal{T}$ used for evaluation. For posts containing each target $t\in \mathcal{T}$, the overall group-level HSD classification performance can be measured by a specific accuracy metric. Then, fairness across targeted groups is reflected by the performance discrepancy among all $|\mathcal{T}|$ targets. Intuitively, the smaller the target-wise discrepancy, the fairer the HSD classifier. In our work, 
to quantify the level of algorithmic fairness, we adopt the well-established notion of \textit{False Positive/Negative Equality Difference} \cite{dixon2018measuring} suggested by Google Jigsaw, abbreviated as FPED/FNED. The only slight difference is that we normalize their values with the number of targets involved to make them comparable across different datasets, and the normalized versions are termed nFPED and nFNED:
    \begin{equation}\label{eq:fairness_metrics}
    \begin{split}
    	\text{nFPED}=\frac{1}{|\mathcal{T}|}\sum_{t\in \mathcal{T}}|\text{FPR}-\text{FPR}(t)|,\\
    	\text{nFNED}=\frac{1}{|\mathcal{T}|}\sum_{t\in \mathcal{T}}|\text{FNR}-\text{FNR}(t)|,
    \end{split}
    \end{equation}
    where $\text{FPR}(t)$ and $\text{FNR}(t)$ are respectively the  false positive and false negative rates on posts associated with target $t$, while FPR and FNR respectively denote the overall false positive and false negative rates on all posts. Essentially, both nFPED and nFNED measure how far on average the HSD performance on each target $t$ deviates from the global average, and ideally will converge to zero if all targets are treated evenly by the HSD classifier. Since it is equally important to account for both nFPED and nFNED for fairness, we further add the harmonic mean of both metrics, termed \textit{harmonic fairness} (HF):
    \begin{equation}\label{eq:HF}
        \text{HF}=2\times \frac{\text{nFPED} \times \text{nFNED}}{\text{nFPED} + \text{nFNED}},
    \end{equation}
which jointly factors in two metrics for fairness evaluation. With the fairness metrics defined, we formulate our HSD task with generalizable target-aware fairness below.

\textit{\textbf{Definition 3}: Hate Speech Detection with Generalizable Target-aware Fairness.} Given a set of posts $\mathcal{S}_{train}=\{(s_i, \mathcal{T}_i, y_i)\}_{i=1}^{|\mathcal{S}_{train}|}$ where the $i$-th post $s_i$ is associated with the binary label $y_i$ and a subset of known targets $\mathcal{T}_i\in \mathcal{T}_{train}$, $\mathcal{T}_i\neq \emptyset$, our objective is to train a debiased HSD classifier $\phi_{\text{HSD}}(\cdot)$. 
The trained classifier $\phi_{\text{HSD}}(\cdot)$ is evaluated on the test set $\mathcal{S}_{test}=\{(s_j, \mathcal{T}_j, y_j)\}_{j=1}^{|\mathcal{S}_{test}|}$, where $\mathcal{T}_j \in \mathcal{T}_{test}$, $\mathcal{T}_j\neq \emptyset$. More importantly, all test-time posts will contain \textit{at least one} target that is \textit{completely unseen} during training, i.e., $ \mathcal{T}_{test}\setminus \mathcal{T}_{train}\neq \emptyset$. Ideally, $\phi_{\text{HSD}}(\cdot)$ is expected to: (1)~provide maximal classification effectiveness, reflected by larger scores in accuracy, F1, and AUC; and (2) yield minimal target-wise bias, reflected by lower scores in nFPED, nFNED, and HF.

\section{GetFair: Model Design}
\begin{figure}[t!]
\centering
\includegraphics[width=3.3in]{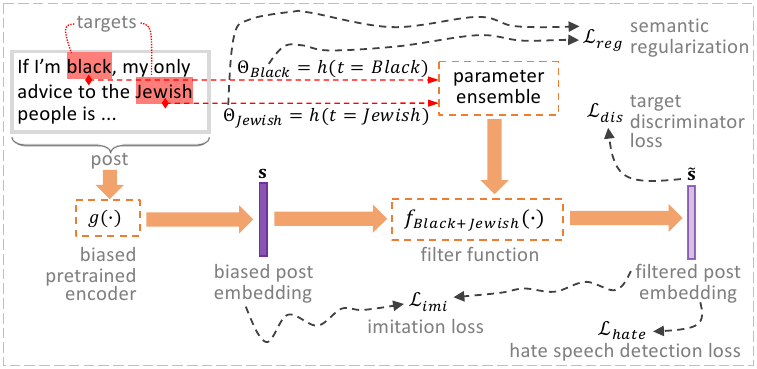}
\vspace{-0.35cm}
\caption{An overarching view of GetFair. Detailed designs of the four objective functions can be found in Sections \ref{sec:L_reg} ($\mathcal{L}_{reg}$), \ref{sec:L_dis} ($\mathcal{L}_{dis}$), and \ref{sec:L_hate_imi} ($\mathcal{L}_{hate}$ and $\mathcal{L}_{imi}$), respectively.}
\label{Figure:GetFair}
\vspace{0.2cm}
\end{figure}

We provide a graphical view of GetFair's workflow in Figure \ref{Figure:GetFair}. With our research task defined, we unfold the design of each core component of it.

\subsection{Target-specific Filter Generation with Adaptive Hypernetwork}\label{sec:filter}
In GetFair, we utilize a filter function $f(\cdot)$ which takes the generated $d$-dimensional content embedding $\textbf{s} \in \mathbb{R}^d$ from an arbitrary encoder $g(\cdot)$ as its input, and emits a debiased representation $\tilde{\textbf{s}} = f(\textbf{s}) \in \mathbb{R}^d$. 
The filter $f(\cdot)$ will be trained to remove target-specific information from the content embedding $\textbf{s}$, such that subsequent hate speech predictions made with $\tilde{\textbf{s}}$ are less reliant on it and will exhibit less performance bias on different targets while maintaining accuracy. Without loss of generality, $f(\cdot)$ can be formulated as a multi-layer perceptron (MLP) with weights $\textbf{W} \in \mathbb{R}^{d\times d}$ and biases $\textbf{b} \in \mathbb{R}^{d}$ at all layers. Considering the vast pool of possible targets in online platforms, it is unrealistic to rely on a single MLP's capacity for removing all targets' relevant information from $\textbf{s}$. Thus, a more performant approach is to instantiate one filter function $f_t(\cdot)$ per target $t$. However, this will inevitably lead to difficulties in scaling to numerous targets in reality and generalizing to unseen targets (e.g., an emerging buzzword) during inference. 

To ensure the practicality of GetFair as a plug-in debiasing method for HSD, we propose an efficient alternative for building target-aware filter functions. Instead of letting every filter function have a dedicated set of trainable parameters, we leverage a hypernetwork \cite{von2020continual} $h(\cdot)$ that can adaptively parameterize each target-specific MLP filter. Concretely, for every target $t$, its dedicated filter parameters are generated via:
\begin{equation}
	\Theta_t^{(l)} = [\textbf{W}_t^{(l)}, \textbf{b}_t^{(l)}] = h_l(\textbf{t}),
\end{equation}
where $h_l(\cdot)$ is the hypernetwork responsible for generating parameters (i.e., weights and biases) for the $l$-th layer of filter $f_t(\cdot)$, and $[\cdot,\cdot]$ denotes horizontal concatenation. For notation simplicity, we use $\Theta_t^{(l)} \in \mathbb{R}^{d\times (d+1)}$ to denote the concatenated weight and bias at each layer. Following common practices in hypernetworks \cite{yan2022apg,shamsian2021personalized}, $h_l(\cdot)$ firstly outputs a flat vector with $d^2+d$ dimensionality, which is further reshaped into the matrix form of $\Theta_t^{(l)}$. 

\textbf{Target Indicators.} Notably, $h_l(\cdot)$ is shared across all targets, and the generated parameters are solely conditioned on its input $\textbf{t}\in \mathbb{R}^{d'}$, which is the target indicator that informs the hypernetwork of the exact target filter to generate. To facilitate effective information filtering and target-wise generalizability, the designed target indicators should meet two characteristics: (1) they should be distinguishable for different targets such that every target-specific filter maintains uniqueness; and (2) they are capable of representing an arbitrary number of unseen targets without any training. Hence, this eliminates some indicator methods commonly used in hypernetworks such as one-hot encoding and learnable embeddings \cite{von2020continual,yan2022apg,shamsian2021personalized}. In GetFair, we take advantage of the wide availability of pretrained word embeddings, namely the GloVe embeddings \cite{pennington2014glove} for composing the target indicator $\textbf{t}$. We have adopted its $300$-dimensional version, thus $d'=300$. Specifically, for target $t=\{v_1, ..., v_k, ..., v_n\}$ represented as a sequence of $n$ vocabularies, we take the mean of all $n$ corresponding word embeddings $\textbf{v}_k \in \mathbb{R}^{d'}$ as its representation, i.e., $\textbf{t}=\frac{1}{n}\sum_{k=1}^n \textbf{v}_k$. 

\textbf{Low-rank Parameterization.} In the default design of the adaptive hypernetwork, the output space for each hypernetwork is $d\times (d+1)$, which is substantially large (e.g., $16,512$ predictions to make for $d=128$) compared with the input dimensionality. Consequently, this creates a low-dimensional bottleneck where the hypernetwork layers are not sufficiently expressive for the downstream predictions \cite{yang2018breaking,bhatia2015sparse,guo2019breaking}, thus being prone to underfitting. Furthermore, although the target-specific filter parameters are no longer stored owing to the adaptive hypernetwork, they are still required for in-memory forward and backward passes during run time, thus significantly limiting the batch size allowed and negatively impacting the training time and memory efficiency. As such, we adopt a low-rank formulation for the filter parameters:
\begin{equation}
    \Theta_t^{(l)} = \textbf{U}^{(l)}_t \textbf{W}^{(l)}_t \textbf{V}^{(l)}_t,
\end{equation}
where $\textbf{U}^{(l)}_t\in \mathbb{R}^{d\times K}$, $\textbf{W}^{(l)}_t\in \mathbb{R}^{K\times K}$,  $\textbf{V}^{(l)}_t\in \mathbb{R}^{K\times (d+1)}$, and the rank $K\ll d$. With this low-rank parameterization, assuming $d=d'$, the memory cost for an $L$-layer target-specific filter is dramatically reduced from $\mathcal{O}(Ld^2+Ld)$ to $\mathcal{O}(LK^2+2LdK+LK)$, which now conveniently supports parallelized batch computing. 

\textbf{Multi-target Filters.} It is worth mentioning that, in real applications, one post might be attributed to several different targets, calling upon the necessity for a mechanism that can simultaneously filter a set of targets $\mathcal{T}_i$ associated with each post's embedding $\textbf{s}_i$. A straightforward way of dosing so is to remove one target-specific information at a time via $\tilde{\textbf{s}}_{i,t}=f_t(\textbf{s}_i)$ for every $t\in \mathcal{T}_i$, then merge all filtered embeddings of the same post (e.g., via sum pooling \cite{li2021towards}) into $\tilde{\textbf{s}}_i$. 
Though each $\tilde{\textbf{s}}_{i,t}$ removes information related to target $t\in \mathcal{T}_i$, it is not necessarily the case for a different $\tilde{\textbf{s}}_{i,t'}$ where $t'\in \mathcal{T}_i$ and $t'\neq t$, as the associated filter $f_{t'}(\cdot)$ is only dedicated to removing information about $t'$. Consequently, the embedding fusion step may introduce unwanted target-specific information back into  $\tilde{\textbf{s}}_i$. To bypass this potential defect, we design a combinatorial approach to form a multi-target filter via parameter ensemble:
\begin{equation}\label{eq:CF}
    \tilde{\textbf{s}}_i = f_{\mathcal{T}_i}(\textbf{s}_i \,|\, \Theta_{\mathcal{T}_i}), \,\,\, \Theta_{\mathcal{T}_i} = \{\Theta_{\mathcal{T}_i}^{(l)}\}_{l=1}^{L},
\end{equation}
and the $l$-th layer of this multi-target filter $f_{\mathcal{T}_i}(\cdot)$ is parameterized by $\Theta_{\mathcal{T}_i}^{(l)}\in \Theta_{\mathcal{T}_i}$, which is calculated as follows:
\begin{equation}
    \Theta_{\mathcal{T}_i}^{(l)} =\frac{1}{|\mathcal{T}_i|}\sum_{t\in \mathcal{T}_i}h_l(\textbf{t}).
\end{equation}
In a nutshell, given a post embedding $\textbf{s}_i$ and its targets $\mathcal{T}_i$, our combinatorial approach aggregates all target-specific filters' parameters first, and then compute the filtered embedding in one go. As such, we can prevent the resultant post embedding $\tilde{\textbf{s}}_i$ from being contaminated by the late fusion of individually filtered embeddings. 

\subsection{Regularizing Filter Parameters with Semantic Gap Alignment}\label{sec:L_reg} 
Compared with learning multiple filter functions, learning a unified adaptive hypernetwork brings substantially lighter parameterization, but also comes at a risk of overfitting some specific target distributions, e.g., the frequent ones, challenging the generlizability. Hence, the hypernetwork, if trained without regularization, can potentially fail to generalize when generating filter weights for completely unseen targets. To prevent the generated filter parameters from undesired homogeneity and thus impaired utility, we impose an additional regularization on the distribution of the generated filter parameters $\Theta_t^{(l)}$ w.r.t. different targets. 
Specifically, for every pair of targets $t$, $t'$ and their indicators $\textbf{t}$, $\textbf{t}'$, we would like to preserve their semantic distance within the generated filter parameters $\Theta_{t}$, $\Theta_{t'}$ too, i.e., $d(\textbf{t}, \textbf{t}') \approx d(\Theta_{t}^{(l)},\Theta_{t'}^{(l)})$ with a distance metric $d(\cdot,\cdot)$. To bypass the unmatched dimensionality between $\textbf{t}$ and $\Theta_t$, we define a semantic gap alignment scheme as the regularization loss $\mathcal{L}_{reg}$ as follows:
\begin{equation}\label{eq:L_reg}
	\mathcal{L}_{reg} = \sum_{l=1}^{L} \sum_{t,t'\in \mathcal{T}_{train}} \Big{(} \text{cos}(\textbf{t}, \textbf{t}') - \text{cos}(\overline{\Theta}^{(l)}_t, \overline{\Theta}^{(l)}_{t'})\Big{)}^2,
\end{equation}
where $\text{cos}(\cdot, \cdot)$ is the cosine similarity, and $\overline{\Theta} \in \mathbb{R}^{d(d+1)}$ is the flattened version of $\Theta$ in vector format to facilitate cosine similarity comparison between filter parameters. By imposing $\mathcal{L}_{reg}$, any two similar/dissimilar targets (as reflected by their indicators) will be assigned corresponding filter functions that reflect the same level of relatedness, assuring the semantic gaps among targets to be mirrored into the parameter space.

\subsection{Target Discriminator}\label{sec:L_dis}
Ideally, each debiased post embedding $\tilde{\textbf{s}}_i$ no longer carries target-specific information that leads to the hypothetically unfair classification results. To ensure that the target filter removes the target-related information, we propose to take advantage of adversarial training to optimize each filter. Specifically, we set up a target discriminator defined as the following:
\begin{equation}
    \hat{\textbf{p}}_i = \text{sigmoid}(\text{MLP}_{dis}(\tilde{\textbf{s}}_i)),
\end{equation}
where $\text{sigmoid}(\text{MLP}_{dis}(\cdot)):\mathbb{R}^d\rightarrow (0,1)^{|\mathcal{T}_{train}|}$ is an MLP with a sigmoid readout that maps the input $\tilde{\textbf{s}}$ into a $|\mathcal{T}_{train}|$-dimensional output, with each entry representing the probability of having a target associated with the post $\tilde{\textbf{s}}_i$. Note that given the existence of multiple targets in a single post,  $\hat{\textbf{p}}_i$ is essentially a collection of individual binary classification scores instead of a softmax distribution. The optimization of $\text{MLP}_{dis}(\cdot)$ is facilitated by minimizing the point-wise log loss below:
\begin{equation}\label{eq:L_dis}
	\mathcal{L}_{dis} = - \sum_{\forall s_i \in \mathcal{S}_{train}} \Big{(}\textbf{p}_i^{\top} \log \hat{\textbf{p}}_i + (1-\textbf{p}_i)^{\top} \log (1-\hat{\textbf{p}}_i)\Big{)},
\end{equation}
where $\textbf{p}_i\in \{0,1\}^{|\mathcal{T}_{train}|}$ is the ground truth multi-hot label derived from every $\mathcal{T}_i$. We will defer the introduction of the entire adversarial training procedure to Section~\ref{sec:adversarial_opt}.

\subsection{Hate Speech Classifier}\label{sec:L_hate_imi}
As the debiasing objective is designed in parallel to the pure HSD task, we hereby define the classifier used for HSD. Simply put, the hate speech detector inherits the traditional binary classifier formulation in most existing HSD methods with its input updated from the biased post embedding $\textbf{s}_i$ to the filtered one $\tilde{\textbf{s}}_i$:
\begin{equation}\label{eq:L_hate}
	\hat{{y}}_i = \text{sigmoid}(\text{MLP}_{hate}(\tilde{\textbf{s}}_i)),
\end{equation}
where $\text{sigmoid}(\text{MLP}_{hate}(\cdot)):\mathbb{R}^d\rightarrow (0,1)$ produces a probability scalar $\hat{{y}}_i$ indicating the likelihood of $\tilde{\textbf{s}}_i$ being a hateful post. Being a binary classification task, log loss is adopted for training:
\begin{equation}
	\mathcal{L}_{hate}=\sum_{\forall s_i \in \mathcal{S}_{train}} \Big{(}y_i \log \hat{y}_i + (1-y_i) \log \hat{y}_i\Big{)},
\end{equation}
where $y_i\in \{0,1\}$ is the binary label of each training instance.

\textbf{Enhancing HSD Performance via Imitation Learning.} To prevent the filtered post embedding $\tilde{\textbf{s}}$ from the byproduct of performance degradation, we further enhance the training of the HSD classifier with imitation learning. Specifically, we ask $\text{MLP}_{hate}(\tilde{\textbf{s}}_i)$ to mimic its response when given the unfiltered version of the post embedding $\textbf{s}_i$. This is achieved by aligning the predicted probability distributions over the two binary classes:
\begin{equation}\label{eq:L_imi}
	\mathcal{L}_{imi}= D_{KL}(\,[\hat{y}_{i},1-\hat{y}_{i}]\,\,||\,\,[\hat{y}_{i}',1-\hat{y}_{i}']\,),
\end{equation}
where $D_{KL}(\cdot||\cdot)$ measures the Kullback-Leibler (KL) divergence between two distributions, and $\hat{y}_{i}' = \text{sigmoid}(\text{MLP}_{hate}(\textbf{s}_i))$ is the prediction based on the unfiltered post embedding $\textbf{s}_i$. The imitation loss $\mathcal{L}_{imi}$ encourages the same hate speech classifier to emit similar decisions no matter whether the post embedding is filtered or not, thus further decorrelating the HSD classifier with the spurious target-related features in the texts.

\subsection{Adversarial Optimization via Alternation}\label{sec:adversarial_opt}
Alongside the fundamental goal of achieving accurate HSD, GetFair encompasses two additional components that have adversarial goals. The target discriminator is designed to identify associated targets for each given post embedding, while the filter function $f(\cdot)$ essentially tries to deceive it such that relevant targets cannot be confidently inferred from the debiased embedding $\tilde{\textbf{s}}_i$. As such, this naturally translates to an adversarial training paradigm where the filter function and target discriminator are mutual adversaries to each other. 

To facilitate the optimization of GetFair, we put forward an alternating training paradigm with Algorithm \ref{alg:train}. Specifically, in the first main loop (lines 5-10), mini-batch gradient descent is performed for the target discriminator, which is trained to infer the target labels from the given post embeddings. In the second main loop (lines 11-16), with the discriminator's parameters frozen, the filter function, along with the HSD classifier, are jointly trained with a synergic loss (line 16) that aims to magnify the target classification error while minimizing other loss terms. It is worth noting that, the pretrained transformer-based encoder $g(\cdot)$ for generating $\textbf{s}$ will also be finetuned throughout this adversarial training procedure. 

\begin{algorithm}[!t]
\begin{spacing}{0.9}
\small
\caption{Optimization Procedure of GetFair}\label{alg:train}
\begin{algorithmic}[1]
\State \textbf{Input:} $\mathcal{T}_{train}$, parameters $\Theta_{enc}$, $\Theta_{hyper}$, $\Theta_{dis}$, $\Theta_{hate}$ respectively  
\Statex \hspace{0.8cm} for the pretrained encoder, hypernetwork, target discriminator, 
\Statex \hspace{0.8cm} and HSD classifier, epoch numbers $N$ and $N'$, loss coefficients 
\Statex \hspace{0.8cm} $\mu$, $\gamma$, and $\lambda$
\State \textbf{Output:} Optimized $\Theta_{enc}$, $\Theta_{hyper}$, $\Theta_{dis}$, $\Theta_{hate}$
\State Randomly initialize $\Theta_{hyper}$, $\Theta_{dis}$, $\Theta_{hate}$;
\Repeat
\For{$epoch_{d}=1,\cdots,K$}
\State Draw a mini-batch from $\mathcal{S}_{train}$;
\State Freeze $\Theta_{hyper}$, $\Theta_{hate}$ and $\Theta_{enc}$;
\State $\tilde{\textbf{s}}\leftarrow$ Eq.(\ref{eq:CF});
\State $\mathcal{L}_{dis} \leftarrow$ Eq.(\ref{eq:L_dis});
\State Update $\Theta_{dis}$ w.r.t. $L_{dis}$;
\EndFor
\For{$epoch_{f}=1,\cdots,K'$}
\State Freeze $\Theta_{dis}$;
\State Draw a mini-batch from $\mathcal{S}_{train}$;
\State $\tilde{\textbf{s}}\leftarrow$ Eq.(\ref{eq:CF});
\State $\mathcal{L}_{reg}\leftarrow$ Eq.(\ref{eq:L_reg}), $\mathcal{L}_{dis} \leftarrow$ Eq.(\ref{eq:L_dis}), 
\Statex \hspace{0.8cm} $\mathcal{L}_{hate}\leftarrow$ Eq.(\ref{eq:L_hate}), $\mathcal{L}_{imi}\leftarrow$ Eq.(\ref{eq:L_imi});
\State Update $\Theta_{hyper}$, $\Theta_{hate}$ and $\Theta_{enc}$ w.r.t. $\mathcal{L}_{hate} + \mu \mathcal{L}_{reg} $ 
\Statex \hspace{0.8cm} $+ \gamma \mathcal{L}_{imi} - \lambda \mathcal{L}_{dis}$;
\EndFor
\Until convergence
\end{algorithmic}
\end{spacing}
\end{algorithm}

\section{Experiments}\label{sec:exp}
To evaluate the efficacy of GetFair in providing accurate yet fair HSD results, we conduct experiments to answer the following research questions (RQs):
\begin{itemize}
	\item[\textbf{RQ1:}] Can GetFair  generalize to unseen target groups and outperform state-of-the-art baselines?
	\item[\textbf{RQ2:}] What is the contribution from each core component of it?
	\item[\textbf{RQ3:}] How sensitive GetFair is to its key hyperparameters?
    \item[\textbf{RQ4:}] Is GetFair compatible to different pretrained text encoders?
\end{itemize}

\begin{table*}[t!]
\caption{Hate speech detection and target-aware fairness results with unseen targets. Numbers in bold face are the best results for corresponding metrics, and the second-best results are underlined. Note that we use ``$\uparrow$'' and ``$\downarrow$'' to indicate higher-is-better and lower-is-better metrics, respectively.}
\vspace{-0.4cm}
\centering
\setlength\tabcolsep{2.5pt}
  \begin{tabular}{c|c|ccc|ccc||ccc|ccc}
    \hline
     \multirow{3}{*}{Dataset} &\multirow{3}{*}{Method} & \multicolumn{6}{c||}{Setting 1} & \multicolumn{6}{c}{Setting 2}\\
     \cline{3-14}
     & & \multicolumn{3}{c|}{Effectiveness ($\uparrow$)} & \multicolumn{3}{c||}{Fairness ($\downarrow$)} & \multicolumn{3}{c|}{Effectiveness ($\uparrow$)} & \multicolumn{3}{c}{Fairness ($\downarrow$)}\\
     \cline{3-14}
    & & Accuracy & F1 & AUC & nFPED & nFNED & HF & Accuracy & F1 & AUC & nFPED & nFNED & HF \\
    \hline
    \multirow{6}{*}{Jigsaw} & THSD \cite{shah2021reducing} & 0.5947 & 0.3581 & 0.5930 & 0.0074 & 0.0537 & 0.0129 & 0.6591 & 0.5137 & 0.6581 & \underline{0.0013} & 0.0838 & \underline{0.0026} \\
    & FairReprogram \cite{zhang2022fairness} & \underline{0.6889} & 0.5961 & \underline{0.8040} & 0.0176 & 0.0563 & 0.0268 & \underline{0.7098} & \underline{0.6157} & \underline{0.7098} & 0.0046 & 0.0878 & 0.0087 \\
    & SAM \cite{ramponi2022features} & 0.6385 & \underline{0.6850} & 0.5971 & 0.1300 & 0.1354 & 0.1326 & 0.6167 & 0.5148 & 0.6160 & 0.1614 & 0.1538 & 0.1575 \\
    & LWBC \cite{kim2022learning} & 0.5069 & 0.6446 & 0.5088 & 0.0138 & \underline{0.0207} & 0.0166 & 0.4348 & 0.5231 & 0.4354 & 0.0202 & \underline{0.0671} & 0.0311 \\
    & FEAG \cite{bansal2023controlling} & 0.6235 & 0.4162 & 0.6358 & \textbf{0.0027} & 0.0358 & \underline{0.0051} & 0.6431 & 0.4591 & 0.6421 & 0.0060 & 0.0861 & 0.0112 \\
     \cline{2-14}
    & \textbf{GetFair}& \textbf{0.7367} & \textbf{0.7262} & \textbf{0.8141} & \underline{0.0028} & \textbf{0.0087} & \textbf{0.0042} & \textbf{0.7777} & \textbf{0.7719} & \textbf{0.8598} & \textbf{0.0011} & \textbf{0.0569} & \textbf{0.0023} \\
    \hline
    \hline
    \multirow{6}{*}{MHS} & THSD \cite{shah2021reducing} & 0.6315 & 0.5120 & 0.6321 & \underline{0.0071} & 0.0120 & 0.0089 & 0.6422 & 0.5575 & 0.6406 & 0.0340 & 0.0515 & 0.0410 \\
    & FairReprogram \cite{zhang2022fairness} & 0.6472 & 0.5220 & \underline{0.8134} & 0.0401 & 0.0978 & 0.0569 & \underline{0.7056} & \underline{0.7123} & \textbf{0.7780} & \underline{0.0245} & \underline{0.0118} & \underline{0.0159} \\
    & SAM \cite{ramponi2022features} & \underline{0.7052} & 0.6581 & 0.6540 & 0.1727 & 0.2326 & 0.1982 & 0.6351 & 0.4642 & 0.6334 & 0.1309 & 0.1406 & 0.1356 \\
    & LWBC \cite{kim2022learning} & 0.5049 & \textbf{0.6688} & 0.6918 & \textbf{0.0064} & \textbf{0.0036} & \textbf{0.0046} & 0.5957 & 0.6424 & 0.5969 & 0.1079 & 0.0996 & 0.1036 \\
    & FEAG \cite{bansal2023controlling} & 0.5923 & 0.3671 & 0.6416 & 0.0201 & 0.0530 & 0.0292 & 0.6691 & 0.6170 & 0.4223 & 0.0511 & 0.0399 & 0.0448 \\
     \cline{2-14}
    & \textbf{GetFair}& \textbf{0.7112} & \underline{0.6592} & \textbf{0.8256} & 0.0081 & \underline{0.0041} & \underline{0.0055} & \textbf{0.7066} & \textbf{0.7302} & \underline{0.7729} & \textbf{0.0019} & \textbf{0.0124} & \textbf{0.0033} \\
    \hline
    \end{tabular}
\label{table:unseen_topic}
\end{table*}

\subsection{Evaluation Datasets}\label{sec:data}
In our experiments, two publicly acquired benchmarks are in use, namely Jigsaw and MHS. To center our test around generalizability to unseen targets, we hold out two targeted groups for evaluation and use the rest for training and validation. For a thorough comparison, we have created \textit{two test settings} on both datasets by choosing different seen and unseen targets. In what follows, we provide a brief overview of both datasets and their target settings below.

\textbf{Jigsaw.} This dataset was released by Google's online content moderation arm, namely Jigsaw\footnote{https://www.kaggle.com/competitions/jigsaw-unintended-bias-in-toxicity-classification/data} to encourage relevant research. A subset of the posts have been tagged with a variety of identity attributes (i.e., targets) based on the identities mentioned in each post. For our evaluation, we collate all posts with target tags. The target allocation for two test settings are as follows:
\begin{itemize}
    \item \textbf{Setting 1:} Seen targets (training and validation) -- \{\textit{Male, Female, Homosexual, Christian, Jewish, Black, Mental Illness}\}; Unseen targets (test) -- \{\textit{Muslim, White}\}
    \item \textbf{Setting 2:} Seen targets -- \{\textit{Male, Homosexual, Christian, Jewish, Mental Illness, Muslim, White}\}; Unseen targets -- \{\textit{Female, Black}\}
\end{itemize}

\textbf{MHS.} The measuring hate speech (MHS) corpus was created and released by \cite{kennedy2020constructing}. MHS combines social media posts from YouTube, Twitter, and Reddit that are manually labelled by crowdsource workers from Amazon Mechanical Turk (AMT). The AMT annotators have also marked the targets associated with each post. We also test GetFair's generalization to unseen targets with two settings:
\begin{itemize}
    \item \textbf{Setting 1:} Seen targets -- \{\textit{Race, Origin, Age, Gender, Disability}\}; Unseen targets -- \{\textit{Religion, Sexuality}\}
    \item \textbf{Setting 2:} Seen targets -- \{\textit{Race, Origin, Gender, Religion, Sexuality}\}; Unseen targets -- \{\textit{Age, Disability}\}
\end{itemize}

Given the space limit, the main statistics and other preprocessing steps are listed in Appendix \ref{app:data}.  

\subsection{Baselines and Metrics}\label{sec:baseline}
To test the effectiveness of GetFair, we compare it with five fairness-aware baseline methods that aim to debias HSD classifiers, namely THSD \cite{shah2021reducing}, FairReprogram \cite{zhang2022fairness}, SAM \cite{ramponi2022features}, LWBC \cite{kim2022learning}, and FEAG \cite{bansal2023controlling}. We defer their details to Appendix \ref{app:baseline}.

\begin{figure}[t!]
\centering
\begin{tabular}{|cc|}
\hline
	\hspace{-0.1cm}\includegraphics[width=2.5in]{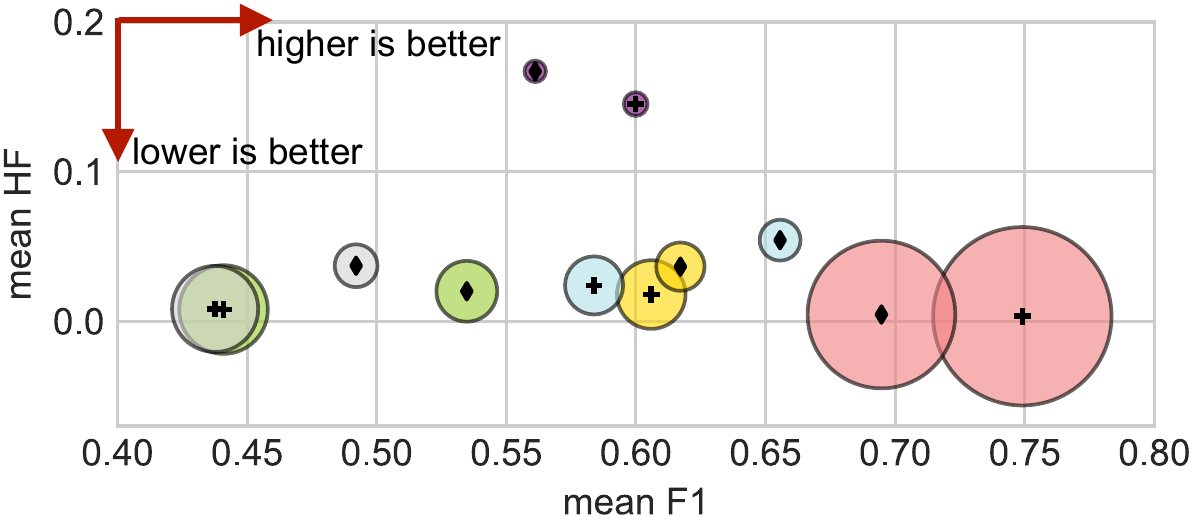}
	&\hspace{-0.35cm}\includegraphics[width=0.72in]{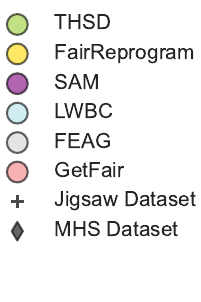}\\
\hline
		\end{tabular}
\vspace{-0.35cm}
\caption{Overall performance visualization with both effectiveness and fairness considered. For each dataset, the mean performance is the average of both settings. The size of each scattered point is proportional to $\frac{mean\,\, F1}{mean\,\, HF}$.}
\label{Figure:ACC_HF}
\vspace{0.4cm}
\end{figure}

\textbf{Metrics.} For evaluation, we cross-compare all methods from two perspectives: (1) the HSD effectiveness, measured by classification metrics accuracy, F1, and Area under the ROC Curve (AUC); (2) the target-specific fairness, measured by nFPED, nFNED, and HF as in Eq.(\ref{eq:fairness_metrics}) and Eq.(\ref{eq:HF}).

\begin{figure*}[t!]
\centering
\begin{tabular}{cccc}
	\hspace{-0.2cm}\includegraphics[width=1.7in]{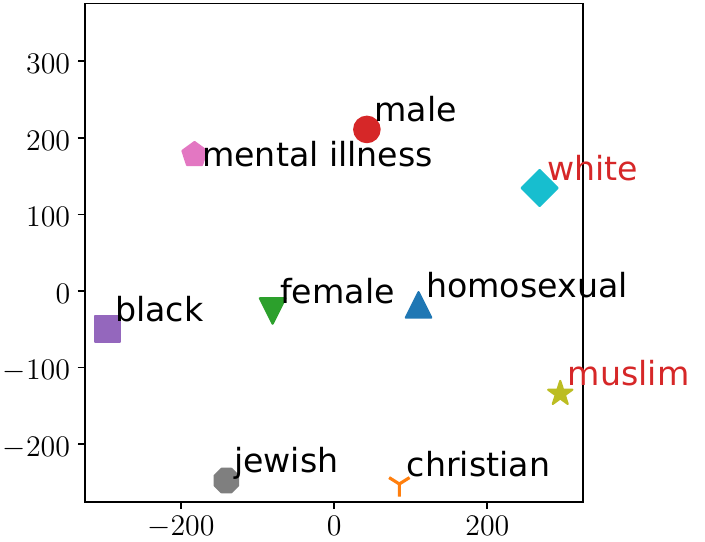}
	&\hspace{-0.2cm}\includegraphics[width=1.68in]{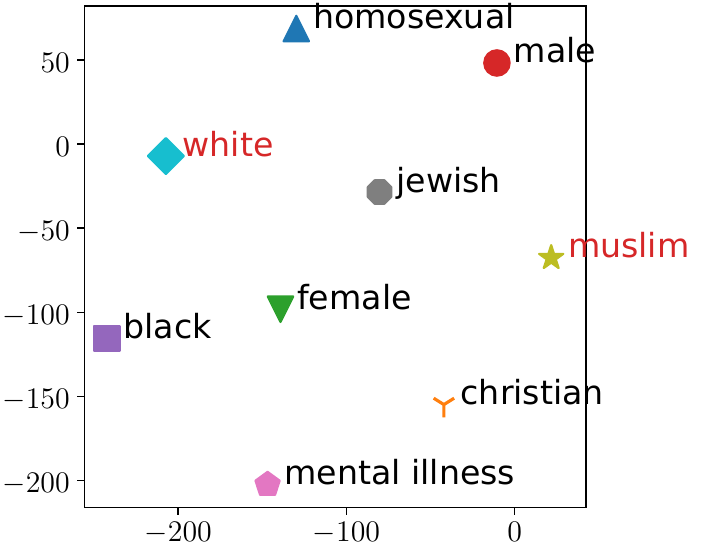}
  	&\hspace{-0.3cm}\includegraphics[width=1.8in]{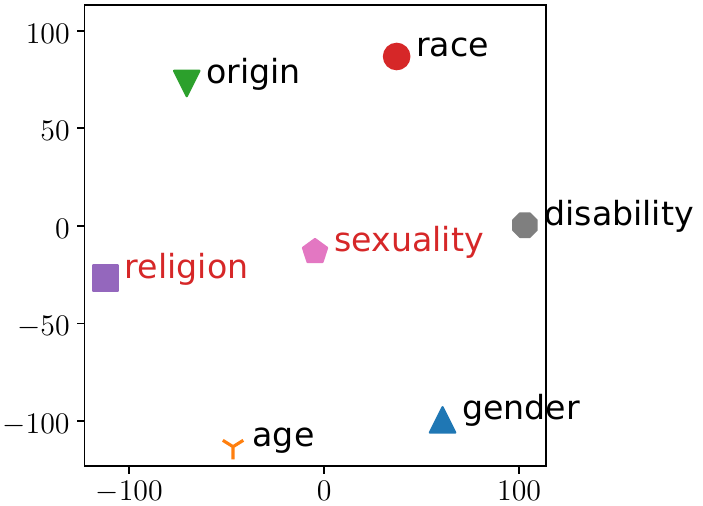}
	&\hspace{-0.4cm}\includegraphics[width=1.7in]{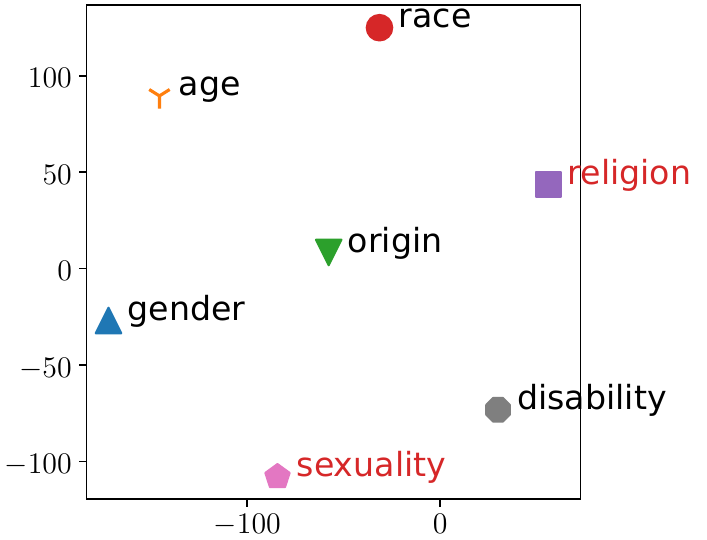}\\
 \hspace{-0.6cm}(a) Targets -- Jigsaw
	&\hspace{-0.6cm}(b) Filters -- Jigsaw
    & \hspace{-0.8cm}(c) Targets -- MHS
	&\hspace{-0.8cm}(d) Filters -- MHS\\
		\end{tabular}
\vspace{-0.35cm}
\caption{The t-SNE \cite{van2008visualizing} visualization of target indicators and the generated target-specific filter parameters. Targets that are only seen in the test set are annotated in red.}
\label{Figure:visualization}
\vspace{-0.4cm}
\end{figure*}

\textbf{Implementation Notes.} For fairness, we adopt the same pretrained encoder, BERT\footnote{We adopt the base version from \url{https://huggingface.co/transformers/v3.1.0/model_doc/bert.html}.}~\cite{kenton2019bert} across all the methods tested for embedding the texts in each post, which is a popular and performant choice in the HSD literature \cite{bhandari2023crisishatemm,gupta2023same}. For GetFair, by default we set discriminator loss coefficient $\lambda=0.9$ and the imitation loss weight $\gamma=3$ on both datasets according to the contributions and magnitudes of the loss terms. Also, we set $\mu$ to $0.9$ and $0.5$, and $K$ to $1$ and $5$ respectively for Jigsaw and MHS. $N$ and $N'$ are set to $1$ and $5$ for optimizing GetFair, respectively. We use hidden dimension $d=256$ on both datasets. The filter depth $L$ is consistently set to $1$ for optimal efficiency as we do not notice a significant performance gain from adding more layers. As for both the target discriminator and HSD classifier, we use the same 3-layer architecture with a consistent $256$ hidden dimension in their MLPs. 

\subsection{Overall Performance (RQ1)}
We start our analysis with the overall performance. Table \ref{table:unseen_topic} reveals the results from all debiasing methods on both datasets. In what follows, we present our observations and findings. 
 
\textbf{Effectiveness in HSD.} Table \ref{table:unseen_topic} has shown the consistently advantageous HSD effectiveness of GetFair, which has achieved the best accuracy, F1, and AUC on Jigsaw, and is either the best or second-best across these three metrics on MHS. In general, the majority of debiased HSD methods achieve the fairness objective by making a considerable sacrifice in the HSD efficacy, witnessed by the subpar accuracy of LWBC on both Jigsaw and MHS, as well as the low F1 of several models, e.g., THSD and FEAG on both Jigsaw and MHS, as well as FairReprogram on MHS. In contrast, on top of its capability of debiasing, GetFair maintains a high level of utility.

\textbf{Fairness in HSD.} On Jigsaw, GetFair yields highly advantageous fairness scores under both settings. Such superiority can also be observed on MHS with Setting 2. On MHS with Setting 1, GetFair has scored the second-best results for both nFNED and HF, only falling behind LWBC by a small margin. In the meantime, it is worth mentioning that the strong fairness of LWBC on MHS is achieved by giving up its utility -- its 50\% accuracy implies the detection effectiveness is only slightly better than a random classifier. It is also noticed that, among the two metrics nFPED and nFNED, some methods are able to obtain a relatively low score on one of them, but tend to get a substantially higher score on the other. Examples include THSD and FEAG on Jigsaw, where the nFNED scores are one magnitude higher than their nFPED scores, translating into a larger number of real hateful posts being missed out by the trained classifier. Meanwhile, GetFair keeps both nFPED and nFNED low, showcasing a balanced performance. More importantly, considering that the test set contains two unseen targets, this reflects the superior generalizability of GetFair in real-world settings. 

\textbf{Performance Summary.} To showcase the effectiveness-fairness trade-off of all methods tested, we visualize their performance on both datasets via a scattered plot in Figure \ref{Figure:ACC_HF}. Based on the visualization, GetFair is in the highest quartile among all the HSD debiasing methods. Again, this verifies that, compared with other baselines, GetFair is able to achieve the state-of-the-art target-aware fairness results without hurting the real-world usability of the trained HSD classifier. As an additional note on efficiency, with a batch size of 128 and a single Nvidia A40 GPU, GetFair consumes 0.1433s and 0.0962s inference time respectively on Jigsaw and MHS, which can fully support real-time detection.

\subsection{Ablation Study (RQ2)}
We hereby answer RQ2 via ablation study, where we build variants of GetFair by removing/modifying one key component at a time, and the new results from the variants are recorded in Table \ref{table:ablation}. We use F1 and HF for performance demonstration, and test with Setting 1 on both datasets. Specifically, we are interested in the contributions of imitation learning, semantic gap alignment, and the design of multi-target filters to both the HSD effectiveness and fairness. In what follows, we introduce the corresponding variants and analyze their performance implications. 

\begin{table}[t]
\caption{Ablation test with different model architectures.}
\vspace{-0.3cm}
\centering
\setlength\tabcolsep{2.5pt}
  \begin{tabular}{c|c|c|c}
    \hline
     Dataset & Architecture & F1 & HF \\
     \hline
    \multirow{4}{*}{Jigsaw} & Default & 0.7262 & 0.0042 \\
    \cline{2-4}
    & Remove Imitation Learning & 0.7189 & 0.0193\\
    & Remove Semantic Gap Alignment & 0.7177 & 0.0432 \\
    & Combinatorial Embedding Filters & 0.3853 & 0.0284 \\
    \hline
    \hline
    \multirow{4}{*}{MHS} & Default & 0.6592 & 0.0055 \\
    \cline{2-4}
    & Remove Imitation Learning & 0.6439 & 0.0051 \\
    & Remove Semantic Gap Alignment & 0.6706 & 0.0152 \\
    & Combinatorial Embedding Filters & 0.4866 & 0.0077 \\
    \hline   
    \end{tabular}
\label{table:ablation}
\vspace{0.1cm}
\end{table}

\textbf{Remove Imitation Learning.} The imitation learning objective defined in Eq.(\ref{eq:L_imi}) aims to uplift the HSD classifier's performance by aligning the predicted probability distributions generated from the same post's filtered and unfiltered embeddings. After removing this component, GetFair has experienced a slight drop in F1 scores on both datasets, which showcases the efficacy of imitation learning for improving the detection accuracy. It is also noticed that while the HF score stays stable on MHS, there is an increase in HF on the Jigsaw dataset. One possible reason is that, the performance decrease has also amplified the model's tendency of producing false negative and false positive results, thus bumping up the HF score. 

\begin{figure*}[t!]
\centering
\begin{tabular}{cccc}
 \multicolumn{4}{c}{\vspace{-0.2cm}\includegraphics[width=1.8in]{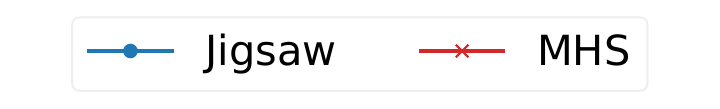}}\\
	\vspace{-0.25cm}\hspace{-0.15cm}\includegraphics[width=1.8in]{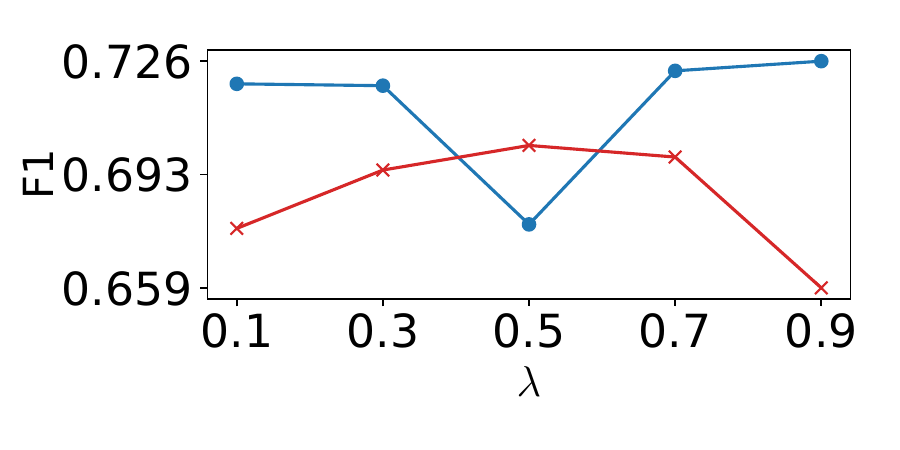}
	&\hspace{-0.5cm}\includegraphics[width=1.8in]{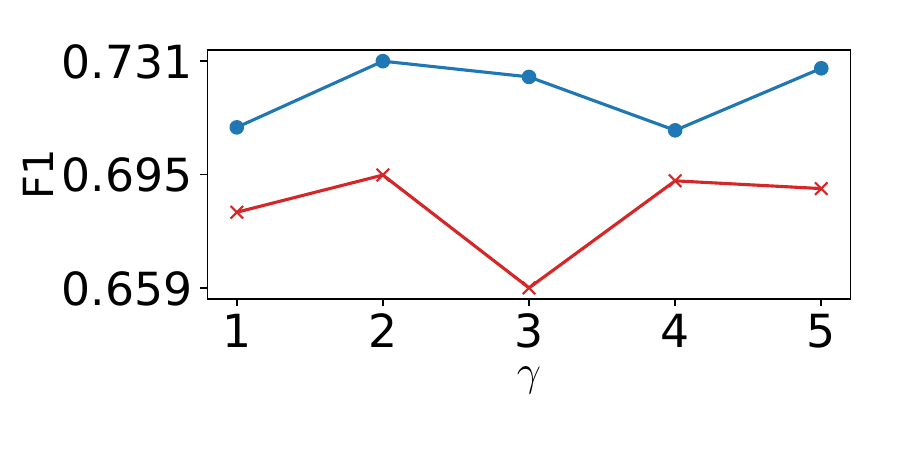}
	&\hspace{-0.5cm}\includegraphics[width=1.8in]{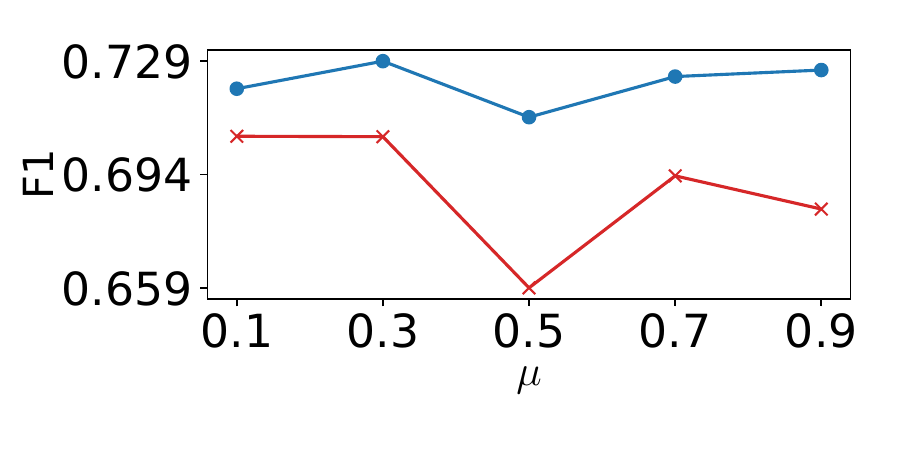}
	&\hspace{-0.5cm}\includegraphics[width=1.8in]{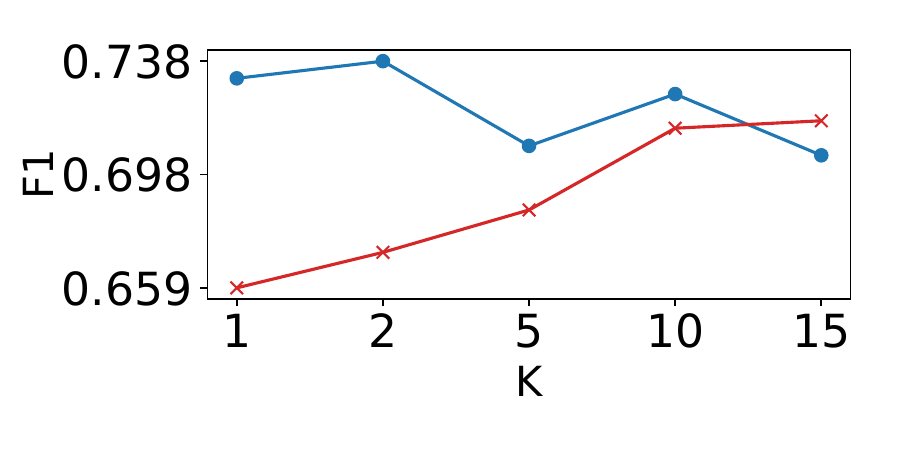}\\
    \hspace{0.35cm}(a) Impact of $\lambda$ on F1
	&\hspace{0.15cm}(b) Impact of $\gamma$ on F1
	&\hspace{0.15cm}(c) Impact of $\mu$ on F1
	&\hspace{0.15cm}(d) Impact of $K$ on F1\\
  	\vspace{-0.25cm}\hspace{-0.05cm}\includegraphics[width=1.8in]{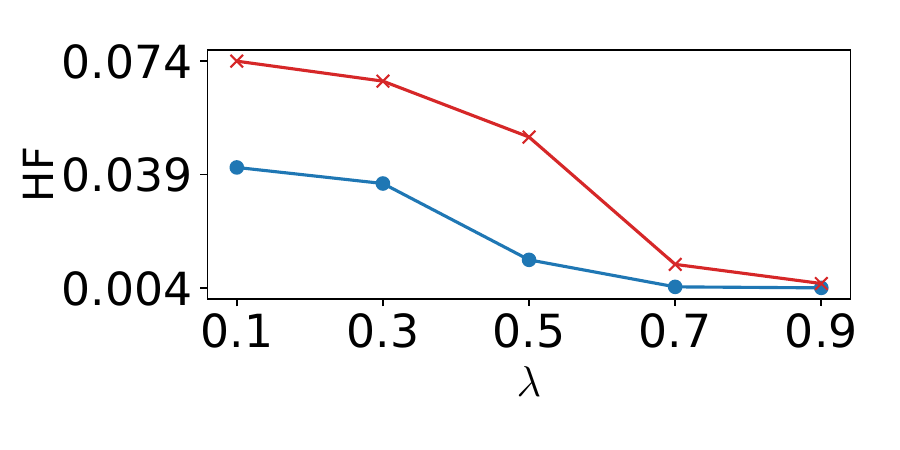}
	&\hspace{-0.5cm}\includegraphics[width=1.8in]{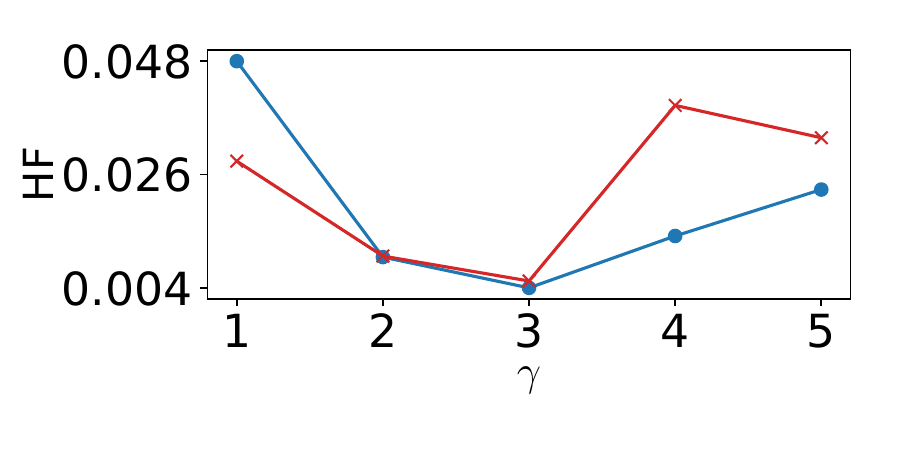}
	&\hspace{-0.5cm}\includegraphics[width=1.8in]{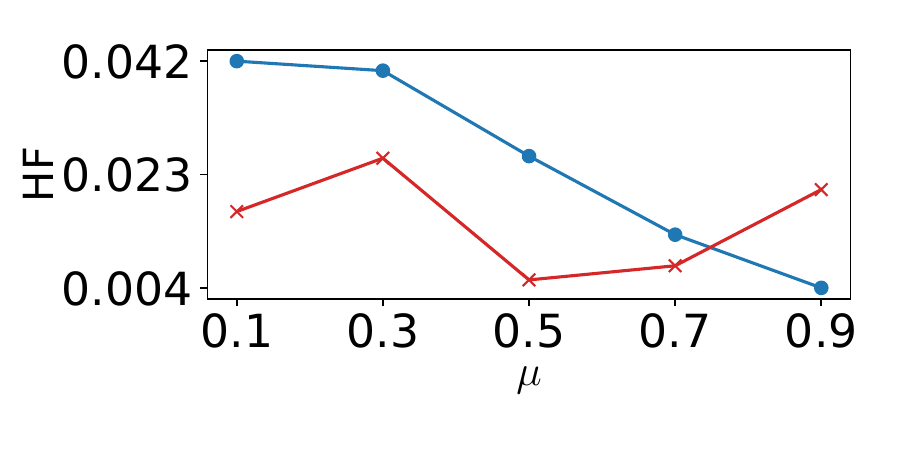}
	&\hspace{-0.5cm}\includegraphics[width=1.8in]{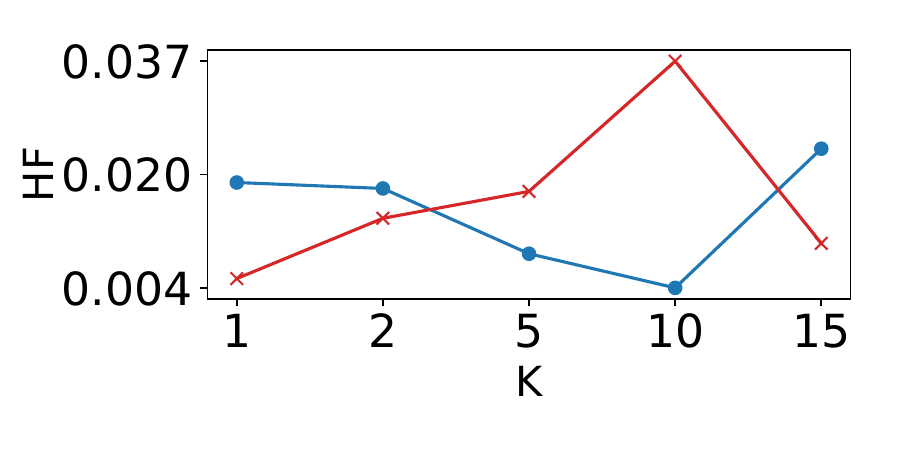}\\
     \hspace{0.4cm}\hspace{-0.15cm}(e) Impact of $\lambda$ on HF
	&\hspace{0.15cm}(f) Impact of $\gamma$ on HF
	&\hspace{0.15cm}(g) Impact of $\mu$ on HF
	&\hspace{0.15cm}(h) Impact of $K$ on HF\\
		\end{tabular}
\vspace{-0.35cm}
\caption{Analysis of the impact from key hyperparameters, with effectiveness and fairness metrics F1 and HF, respectively.}
\label{Figure:group_size}
\vspace{-0.4cm}
\end{figure*}

\textbf{Remove Semantic Gap Alignment.} With the semantic gap alignment regularizer (Eq.(\ref{eq:L_reg})) removed, the fairness of the HSD results has significantly deteriorated. As the regularization essentially rectifies the way a target-specific filter's parameters are generated, its removal has incurred a lower quality of generated filters and consequently inferior generalizability to unseen targets. On MHS, a small improvement of F1 is observed, which may attribute to the use of target-specific features as a result of less effective filtering. Besides, to further showcase the efficacy of the semantic gap alignment, we have visualized both the target indicators and their corresponding filter weights generated by the hypernetwork (with the semantic gap alignment in place) by projecting them onto a two-dimensional space via t-SNE \cite{van2008visualizing} via Figure \ref{Figure:visualization}. As can be seen, the distances among the target indicators (i.e., raw inputs of the hypernetwork) are mostly kept in their corresponding filter parameters (i.e., outputs of the hypernetwork), thus explaining the fairness boost from this semantic gap alignment scheme. 

\textbf{Replacing Parameter Ensemble with Combinatorial Embedding Filters.} As we put forward a filter design explicitly for the co-existence of multiple targets within one post, we compare its performance with a straightforward counterpart, i.e., simply performing sum pooling on each individually filtered post embeddings as described in Section \ref{sec:filter}. That is to say, Eq.(\ref{eq:CF}) is updated to $\tilde{\textbf{s}}_i =\frac{1}{|\mathcal{T}_i|}\sum_{t\in \mathcal{T}_i}\tilde{\textbf{s}}_{i,t}$. With this change, a dramatic drop in both the HSD accuracy and fairness is observed. This has verified that, compared with combining outputs from all single-target filters, our parameter ensemble design in GetFair is more capable of preventing the filtered post embedding from noises.

\subsection{Hyperparameter Sensitivity (RQ3)}

In this section, we examine the impact of some core hyperparameters, namely the discriminator weight $\lambda$, imitation weight $\gamma$, and regularization weight $\mu$ used by the synergic loss in Algorithm \ref{alg:train}, as well as the rank $K$ (Section \ref{sec:filter}). This is done by adjusting one hyperparameter at a time and recording the new results achieved, while all other hyperparameters are kept to the default setting. This part of the experiments are also conducted with Setting 1 on both datasets, and similar trends can be observed with Setting 2. 

\textbf{Impact of $\lambda$.} The coefficient $\lambda$ is tuned in $\{0.1, 0.3, 0.5, 0.7, 0.9\}$. As it essentially links to the optimization of target-specific filtering, $\lambda$ has a stronger impact on HF than on F1, where HF benefits from a larger value of $\lambda$. When $\lambda$ increases from $0.1$ to $0.5$, there is a rapid improvement in the classification fairness as per the HF score. When $\lambda$ is sufficiently large (i.e., $\lambda\geq 0.7$ in our case), the fairness gain remains positive but appears to be at a lower rate. 

\textbf{Impact of $\gamma$.} Interestingly, the variation of imitation loss weight $\gamma$ not only causes fluctuations in the HSD effectiveness, but also correlates to different fairness levels. As GetFair primarily puts more value on fairness in the HSD, $\gamma=3$ is a reasonable choice for a balanced fairness-accuracy trade-off. 

\textbf{Impact of $\mu$.} The regularization weight controls the diversity and quality of the hypernetwork-generated target filters, thus higher sensitivity to $\mu$ is observed on the fairness metric HF. The general trend is that, as $\mu$ grows, fairer detection outcomes are attained. When taking the corresponding F1 into account, $\mu=0.9$ and $\mu=0.5$ are respectively the most sensible settings for Jigsaw and MHS. 

\begin{table}[t!]
\caption{Performance of GetFair when paired with different pretrained text encoders.}
\vspace{-0.3cm}
\centering
\setlength\tabcolsep{5.5pt}
  \begin{tabular}{c|c|c|c}
    \hline
     Dataset & Pretrained Encoder $g(\cdot)$ & F1 & HF \\
     \hline
    \multirow{3}{*}{Jigsaw} & BERT-base (Default) & 0.7262 & 0.0042\\
    \cline{2-4}
    & DistilGPT2 & 0.6988 & 0.0051 \\
    & RoBERTa-base & 0.7501 & 0.0071 \\
    \hline
    \hline
    \multirow{3}{*}{MHS} & BERT-base (Default) & 0.6592 & 0.0055 \\
    \cline{2-4}
    & DistilGPT2 & 0.6879 & 0.0055 \\
    & RoBERTa-base & 0.7864 & 0.0034 \\
    \hline   
    \end{tabular}
\label{table:otherPLMs}
\vspace{0.1cm}
\end{table}

\textbf{Impact of $K$.} The low-rank parameterization of the filter parameters generated by the hypernetwork intends to lower the memory cost while ensuring an adequate level of efficacy. Intuitively, a larger $K$ is able to preserve more expressiveness for the generated filters. On Jigsaw, a larger $K$ generally contributes to lower F1 but better HF, while the trend on MHS is on the opposite side. A possible reason is that, on Jigsaw, a stronger filter function removes more target-related information from the post embedding, but also blocks useful features for HSD classification; while on MHS, the spurious target-related features are more implicit, hence a more capable filter function can effectively filter out those noises from the post embedding to achieve a higher F1 score but also more false positive/negative predictions (i.e., a higher HF).

\subsection{Compatibility with Other Encoders (RQ4)}
GetFair is designed to be compatible with different pretrained text encoders $g(\cdot)$ as backbones. To verify this compatibility, we have tested GetFair by using two other popular pretrained language models (PLMs), namely DistilGPT2 \cite{huggingface2019distilgpt2} and RoBERTa (base version) \cite{liu2019roberta} as $g(\cdot)$. Similar to RQ2 and RQ3, we test on both datasets under Setting 1 and report F1 and HF metrics.
 
From the results in Table \ref{table:otherPLMs}, the first conclusion drawn is that GetFair is able to maintain its performance in HSD tasks when it is paired with different pretrained encoders, especially the classification fairness measured by HF. Secondly, as per F1, RoBERTa yields the highest HSD effectiveness among the three backbones, while DistilGPT2 shows limited effectiveness gain compared with the default BERT used in GetFair. We hypothesize that this is attributed to  different PLMs' model capacity. As per the PLMs tested, DistilGPT2 has a lower capacity than the BERT-base we have used (89 million and 110 million parameters respectively), hence producing a similar or even lower accuracy than BERT. In the meantime, the better accuracy of RoBERTa aligns with its higher model size (125 million parameters in the base version) and capacity.

\section{Related Work}\label{sec:related}
In this section, we review the recent advances in fields that are relevant to our work, specifically hate speech detection (HSD) and the fairness aspects of HSD tasks.

\subsection{Hate Speech Detection}
Mining user-generated web content \cite{tam2019anomaly,sun2022structure,nguyen2017argument} is a long-lasting research area with versatile applications \cite{zhang2022pipattack,nguyen2017retaining,chen2020try}, where hate speech detection is one of the most representative lines of work. Hate speech on social media is commonly defined as a language that attacks or diminishes, that incites violence or hate against groups, based on specific characteristics such as physical appearance, religion, gender identity or other \cite{fortuna2018survey}. In summary, extracting representative linguistic features from the post texts lies at the core of various HSD methods. Early practices in HSD involve the use of vocabulary dictionaries like Ortony Lexicon \cite{dinakar2011modeling} to pinpoint potentially hateful keywords, which then evolves to the use of more sophisticated linguistic features like the term frequency with inverse document frequency (TF-IDF) \cite{dinakar2011modeling}, $N$-grams \cite{greevy2004classifying}, and sentiment \cite{nobata2016abusive}. Those text-based features can be easily incorporated with the downstream classifiers for hate speech classification. In the most recent line of work, there has been an adoption of more complex features like images \cite{kiela2020hateful,bhandari2023crisishatemm} and social connections \cite{ribeiro2018characterizing,masud2021hate} for the HSD task. Meanwhile, with the rise of language models, especially variations of the transformer family \cite{vaswani2017attention,sanh2019distilbert,radfordimproving}, 
features extracted from the raw texts (a.k.a. embeddings) by those pretrained language models are arguably the most widely adopted option in HSD \cite{kiela2020hateful,lee2021disentangling,bhandari2023crisishatemm,gupta2023same}. The commonly shared goal of HSD is usually performance-oriented, where many new aspects in HSD are attracting an increasing amount of attention, such as efficiency and timeliness of the predictions \cite{tran2020habertor}, explainability and transparency of the classification mechanism \cite{mathew2021hatexplain,yang2023hare}, and robustness of the HSD classifier to adversarial attacks or low-quality data \cite{sheth2023peace}. In what follows, we discuss an emerging research topic in this area, i.e., ensuring the fairness in HSD. 

\subsection{Fairness in Hate Speech Detection}
A series of research has uncovered that methods in natural language processing (NLP) tasks are subject to a variety of fairness issues \cite{madanagopal2023bias,omrani2023social,ramponi2022features}, and HSD is no exception as a typical NLP task. Technically, in the context of HSD, the biases can come from either the source \cite{badjatiya2019stereotypical,caliskan2017semantics} (e.g., authors, annotators, and data collectors) and the target~\cite{ramponi2022features,shah2021reducing} of online posts. In this work, our main focus is to demote biases toward the targets of online posts. 

The main objective of addressing biases against targeted groups is to diminish the emphasis placed on the inclusion/exclusion of particular terms and redirect attention towards the broader context within the content being evaluated. As discussed in Section \ref{sec:intro}, data-centric solutions are a representative line of work. Some solutions reweigh each training sample based on its likelihood of introducing bias into the model \cite{zhang2020demographics,schuster2019towards,zhou2021challenges}, while some provide additional meta-data, human annotations, or data augmentation \cite{ramponi2022features,dixon2018measuring,kennedy2020contextualizing,sen2022counterfactually} for a more rigorous model training process. To bypass the reliance on empiric and human involvement, model-centric solutions aim at removing information related to the spurious target-related features from learned representations of an online post. This is usually achieved by utilizing a dedicated filter module, which is trained end-to-end along with the HSD classifier \cite{hauzenberger2023modular,gupta2023same,kumar2023parameter,cheng2022bias}. It is also seen that some solutions are able to handle intersectional bias between multiple targets \cite{maheshwari2023fair,tan2019assessing}. However, the aforementioned methods all suffer from restricted generalizability, as they are trained with the assumption that all targets are seen in the training stage. Due to the distributional discrepancies among different targets, the debiasing effectiveness can hardly transfer to completely unseen targets during inference, creating a practicality bottleneck for real-world applications. Despite the efforts on some generalizable HSD methods \cite{bourgeade2023did,ranasinghe2020multilingual}, their goal is to maintain the HSD accuracy under distributional/domain shift, thus being unable to solve the generalizability challenges associated with target-aware fairness.

\section{Conclusion}\label{sec:conclusion}
In this paper, to address the deficiency of existing debiasing/fairness-aware HSD methods when handling unseen targets during training, we propose GetFair, which achieves generalizable target-aware fairness in HSD by adaptively generating target-specific filters via a hypernetwork instead of training individualized ones. A suite of innovative designs including low-rank parameterization, semantic gap alignment, and imitation learning are proposed for lowering the memory cost, regularizing the generalizability of target-specific filters, and enhancing the HSD classifier, respectively. Through a series of experiments, we have validated the effectiveness, fairness, and generalizability of GetFair, proving it to be a viable solution to ensuring target-aware fairness in HSD. Possible extensions of GetFair in our future work include discovering time-sensitive patterns \cite{chen2018tada,chen2020sequence} for dynamic debiasing, as well as developing lightweight variants \cite{yin2024device} of GetFair to further enhance scalability. 

\section*{Acknowledgement}
This work is supported by Australian Research Council under the streams of Discovery Project (Grant No. DP240101108 and DP240101814), Future Fellowship (No. FT210100624), Discovery Early Career Researcher Award (No. DE230101033 and DE220101597), Center of Excellence (No. CE200100025) and Industrial Transformation Training Centre (No. IC200100022). This work is partially supported by the Swiss National Science Foundation (Contract No. CRSII5\_205975). Financial support from The University of Queensland School of Business in the UQBS 2023 Research Project is gratefully acknowledged.

\newpage
\bibliographystyle{ACM-Reference-Format}
\balance


\section*{Appendix}
\appendix

\section{Dataset Statistics}\label{app:data}
As discussed in Section \ref{sec:data}, we provide the detailed statistics of our experimental datasets, namely Jigsaw and MHS. Both datasets are inherently imbalanced, where the non-hateful posts outnumbers hateful ones. Specifically, we have balanced the binary classes of hateful and non-hateful (neutral) posts in the validation and test sets to ensure a more rigorous evaluation. Note that there can be more than one identified target associated with each post. The posts in the test set are allowed to also contain targets that are seen during training, while the training set does not have any posts that mention the two hold-out targets.

Table \ref{table:Dataset_Jigsaw} lists the key statistics of the Jigsaw dataset after being processed for the two different unseen target settings. Analogously, Table \ref{table:Dataset_MHS} below corresponds to the MHS dataset, also with two different unseen target settings.

\section{An Overview of Baselines}\label{app:baseline}
Following Section \ref{sec:baseline}, we hereby provide an extended summary of the baseline methods tested:
\begin{itemize}
	\item \textbf{THSD}: It stands for the token level hate sense disambiguation (THSD) approach proposed by Meta AI Research \cite{shah2021reducing}, which regularizes the HSD via token-level consensus to make the classification rely less on target-specific signals.
	\item \textbf{FairReprogram}: This approach appends to the input a set of learnable perturbations called the fairness trigger \cite{zhang2022fairness} to achieve the fairness objective under a min-max formulation. 
	\item \textbf{SAM}: It is a data-centric approach that utilizes spurious artifact masking (SAM) \cite{ramponi2022features} on the input tokens when training the HSD towards both the accuracy and fairness goals. 
	\item \textbf{LWBC}: An ensemble method named learning with biased committee (LWBC) \cite{kim2022learning} is proposed to adaptively discover and reweight bias-conflicting samples during training. 
	\item \textbf{FEAG}: This is the state-of-the-art debiasing method for text classifiers \cite{bansal2023controlling} that leverages feature effect augmentation (FEAG) to counter the spurious correlations learned.
\end{itemize}

\begin{table}[h!]
\caption{Statistics of the Jigsaw dataset.}
\vspace{-0.5cm}
\setlength\tabcolsep{0.78pt}
\center
  \begin{tabular}{c|c|c|c|c}
    \hline
     \multirow{2}{*}{Setting} & \multirow{2}{*}{Split} & {\#Hateful } & {\#Non-hateful} & \multirow{2}{*}{Targets} \\
      & & Posts & Posts & \\
     \hline
    \multirow{7}{*}{1} & \multirow{5}{*}{Train} & \multirow{5}{*}{11,825} & \multirow{5}{*}{98,396} & \textit{Male, Female,}\\
    &  &  &  & \textit{Homosexual,} \\
    &  &  &  & \textit{Christian,} \\
    &  &  &  & \textit{Jewish, Black,} \\
    &  &  &  & \textit{Mental Illness} \\
    \cline{2-5}
    & Validation & 3,939 & 3,826 & Same as above \\
    \cline{2-5}
    & Test & 5,973 & 6,031 & + \textit{Muslim, White} \\
    \hline
    \hline
     \multirow{7}{*}{2} & \multirow{5}{*}{Train} & \multirow{5}{*}{11,734} & \multirow{5}{*}{76,844} & \textit{Male, Jewish} \\
    &  &  &  & \textit{Homosexual,} \\
    &  &  &  & \textit{Christian,} \\
    &  &  &  & \textit{Muslim, White,} \\
    &  &  &  & \textit{Mental Illness} \\
    \cline{2-5}
    & Validation & 3,764 & 4,104 & Same as above \\
    \cline{2-5}
    & Test & 5,573 & 5,608 & + \textit{Female, Black} \\
    \hline
    \end{tabular}
\label{table:Dataset_Jigsaw}
\end{table}

\begin{table}[h!]
\caption{Statistics of the MHS dataset.}
\vspace{-0.5cm}
\setlength\tabcolsep{0.78pt}
\center
  \begin{tabular}{c|c|c|c|c}
    \hline
     \multirow{2}{*}{Setting} & \multirow{2}{*}{Split} & {\#Hateful } & {\#Non-hateful} & \multirow{2}{*}{Targets} \\
      & & Posts & Posts & \\
      \hline
    \multirow{4}{*}{1} & \multirow{2}{*}{Train} & \multirow{2}{*}{4,118} & \multirow{2}{*}{14,585} & \textit{Race, Origin, Age,} \\
    &  &  &  & \textit{Gender, Disability} \\
    \cline{2-5}
    & Validation & 1,480 & 1,474 & Same as above \\
    \cline{2-5}
    & Test & 1,496 & 1,489 & + \textit{Religion, Sexuality} \\
    \hline
    \hline
    \multirow{4}{*}{2} & \multirow{2}{*}{Train} & \multirow{2}{*}{6,782} & \multirow{2}{*}{24,610} & \textit{Race, Origin, Gender,} \\
    &  &  &  & \textit{Religion, Sexuality} \\
    \cline{2-5}
    & Validation & 2,566 & 2,707 & Same as above \\
    \cline{2-5}
    & Test & 277 & 282 & + \textit{Age, Disability} \\
    \hline
    \end{tabular}
\label{table:Dataset_MHS}
\end{table}

\end{document}